\documentclass{article}

\usepackage{placeins}

\usepackage{subcaption}
\usepackage{arxiv}
\usepackage{tablefootnote}

\usepackage[utf8]{inputenc} 
\usepackage[T1]{fontenc}    
\usepackage{hyperref}       
\usepackage{url}            
\usepackage{booktabs}       
\usepackage{amsfonts}       
\usepackage{nicefrac}       
\usepackage{microtype}      
\usepackage{lipsum}		
\usepackage{graphicx}
\usepackage{doi}
\usepackage{CJKutf8}
\usepackage{multirow}

\title{A Survey of Personality, Persona, and Profile in Conversational Agents and Chatbots}

\author{ \mbox{\href{https://orcid.org/0000-0002-5549-5691}{\includegraphics[scale=0.06]{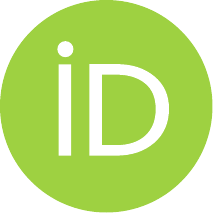}\strut}}\hspace{1mm}Richard Sutcliffe \\
	School of Computer Science and Electronic Engineering \\ University of Essex \\
	Wivenhoe Park, Colchester CO4 3SQ, UK\\
	\texttt{rsutcl@essex.ac.uk} \\
}

\date{}

\renewcommand{\headeright}{R. Sutcliffe}
\renewcommand{\undertitle}{}
\renewcommand{\shorttitle}{Personality, Persona, and Profile in CAs \& Chatbots}

\hypersetup{
pdftitle={A Survey of Personality, Persona, and Profile in Conversational Agents and Chatbots},
pdfsubject={cs.SD, eess.AS, cs.LG},
pdfauthor={Richard Sutcliffe},
pdfkeywords={chatbot, conversational agent, personality, persona, profile, survey, review},
}

\begin{document}
\maketitle

\begin{abstract}
We present a review of personality in neural conversational agents (CAs), also called chatbots. First, we define Personality, Persona, and Profile. We explain all personality schemes which have been used in CAs, and list models under the scheme(s) which they use. Second we describe 21 datasets which have been developed in recent CA personality research. Third, we define the methods used to embody personality in a CA, and review recent models using them. Fourth, we survey some relevant reviews on CAs, personality, and related topics. Finally, we draw conclusions and identify some research challenges for this important emerging field.
\end{abstract}

\keywords{chatbot \and conversation agent \and conversational AI \and neural network \and personalisation \and personality \and persona \and personalised chatbot \and profiling \and review \and survey}

\section{Introduction} \label{introduction}
For a person to converse with a machine was one of the earliest aims of Artificial Intelligence. We define a Conversational Agent (CA) as a piece of software which can interact with a user using natural language \cite{diederich2019towards}. We will commence with a brief history of CAs. In 1962, Austin \cite{austin1962things} was one the first writers to consider what language means in a conversational context rather than just looking at isolated sentences. This line of thinking was continued by Searle \cite{searle1969speech}. In 1966, the ELIZA program of Weizenbaum \cite{weizenbaum1966eliza} used simple string-based techniques to converse with a user. The approach seemed very effective and prompted a general interest in conversational systems. In the 1970s, much interesting work was done. For example, PARRY \cite{colby1971artificial} continued the pattern-matching approach, SHRDLU could converse with a user about a blocks world \cite{winograd1972understanding}, the LUNAR \cite{woods1973progress} could converse about moon rocks.

In the 1980s, deep symbolic approaches continued with systems like BORIS \cite{lehnert1983boris} which could converse in detail about a narrative. In the meantime, pattern-based approaches continued, for example using ALICE \cite{shawar2005chatbot}, which allowed templates to be specified in AIML. In 1999, TREC launched its Question Answering track \cite{voorhees1999trec}. This was a breakthrough, because it showed that real questions could be answered from large-scale newspaper collections. On the conversational front, the pattern-matching approach was refined to one based on recognising intents \cite{liu2019review}. In the 2000s, this data-driven focus led first to conversational systems based on Information Retrieval (retrieve the best response from a large existing set \cite{jafarpour2010filter}, and then systems inspired by statistical machine translation \cite{ritter2011data}.

At the same time, Neural Networks (NNs) were developing rapidly, with the invention of backpropagation and recurrent networks. This led to seq2seq models \cite{sutskever2014sequence} and to the first NN conversation systems \cite{vinyals2015neural,shang2015neural,sordoni2015neural}. From the beginning of CAs, personality was an important factor. For example, ELIZA was a psychoanalyst and PARRY was paranoid. However, with data-driven approaches and NNs, there was potential for much greater progress. In 2016 the first NN paper concerned with personality in CAs appeared, leading to the start of a new era in the development CAs with personality \cite{al2016conversational}.

The aim of this review is to explain and assess progress which has been made in developing CAs with personality. The approach taken is as follows. \textbf{First}, we define Personality, Persona and Profile. We systematically explain and review all personality schemes which have been used in CAs. We also list all models, classified by the personality scheme which they adopt. 
Almost all current work on personality involves training on one or more datasets. Therefore, \textbf{Second}, we describe in detail the various datasets which have been developed in recent CA personality research.
\textbf{Third}, we define the main methods which are used to embody personality in a conversational model. We review works based firstly on the personality scheme, and within that, by the method of embodiment.
\textbf{Fourth}, we survey some relevant reviews on CAs, personality, and related topics. Finally, \textbf{Fifth} we draw some conclusions from this review and highlight some possible areas for future research.

In summary, the contributions of this article are:
\begin{itemize}
\item We provide a systematic survey of personality, persona and profile in Conversational Agent (Chatbot) research;

\item We define and explain all current methods of specifying personality, cross-referenced with all the models discussed;

\item We list the main means by which personality can be embodied in a Conversation Agent, cross-referenced with all the models;

\item We describe in detail 21 datasets used to train CA models with personality;

\item We outline nine topics in personality research, as well as nine further topics related to the research, and list many papers touching upon these topics;

\item We provide all the above information in a series of six tables, as well as discussing it in the text.
\end{itemize}



\section{Personality, Persona, and Profile} \label{personality_defined}
Burger \cite{burger2014personality} describes personality as`consistent behavior patterns and intrapersonal processes originating within the individual' (p4). He then goes on to divide personality research into six approaches, psychoanalytic, trait, biological, humanistic, behavioural/social learning, and cognitive. Lessio and Morris, in the context of Conversational Agents, describe personality as `a set of traits that is stable across situations and time and acts
as a guiding influence on agent behavior and interactions' \cite{lessio2020toward}. Concerning the term Persona, Pradhan and Lazar \cite{pradhan2021hey} combine definitions from Google \cite{google2019create} and Kim et al. \cite{kim2019designing} in order to state: `A persona for a conversation agent is a fictional character and can have a name, age, education or job, or even a defined backstory and personalities'. Finally, a Profile can be defined as a set of pieces of information about a person which can guide the operation and responses of a chatbot.
These definitions are all somewhat imprecise and the distinctions between these three terms vary from paper to paper. Often, authors will use Personality in one paragraph and then Persona for the very same concept in the next. Therefore, in this work, we treat the three terms as interchangable, but use the terminology of the authors where possible. For similar reasons, we will refer to Conversational Agent, Conversational AI, Chatbot and Bot in an interchangable manner, even though distinctions are drawn between them in other works.

In Table \ref{table_personality_method} we can see various definitions of Personality/Persona/Profile which have been used in conversation agent research. For each one, the table gives the seminal paper and a short textual explanation. Note that Table \ref{table_personality_specification} has the same rows as the former table, but lists all the papers which use these personality schemes in their models. In this way, we can see a distribution of models over schemes.

We will run through the personality definitions in Table \ref{table_personality_method}, giving a short history of personality along the way. Where a term is in bold font, it corresponds to a line in the table.


Carl Jung developed a theory of intrapsychic processes called Analytical Psychology \cite{jung1923psychological}. In his approach, a person's preferences and inner desires are examined in a scientific and logical way. Jung contended that there were various dimensions of personality such as conscious versus unconscious, thinking versus feeling, and progression versus regression. Among these, emphasis was placed on \textbf{Introversion versus Extraversion}. On the one hand, there is the preference for concentrating on the inner world of the subjective by means of reflective and introspective activity; this is intraversion. On the other hand there is the preference for concentrating on the outer world of real events together with an active involvement in the environment; this is extraversion.

\begin{table}[t]
\centering
\caption{Methods of specifying personality.}
\begin{tabular}{|p{0.32\linewidth}|p{0.14\linewidth}| p{0.47\linewidth}|} 
\hline 
\textbf{Pre-existing Personality Schemes} & \textbf{Definition} & \textbf{Explanation} \\
\hline

Introversion / Extraversion & \cite{jung1923psychological} & Just these two categories. \\ \hline

Personality Dimensions & \cite{eysenck1953structure} &
Extraversion-Introversion, Neuroticism-Stability and Psychoticism. \\ \hline

Big-5 / OCEAN & \cite{norman1963toward} \cite{norman1966raters} \cite{gosling2003very} & Extroversion, Agreeableness, Openness, Conscientiousness, Neuroticism. \\ \hline

16 Personality Factor (16PF) & \cite{cattell1967functional} \cite{cattell2008sixteen} & Warmth, Reasoning, Emotional Stability, Dominance, Liveliness, Rule-Consciousness, Social Boldness, Sensitivity, Vigilance, Abstractedness, Privateness, Apprehension, Openness to Change, Self-Reliance, Perfectionism, and Tension. \\ \hline

Interpersonal Reactivity Index & \cite{davis1980interpersonal} & Empathy classified on four scales: Perspective Taking, Empathic Concern, Personal Distress, and Fantasy. \\ \hline \hline

\textbf{Chatbot Personality Schemes} & \textbf{Definition} & \textbf{Explanation} \\
\hline

Traits & \cite{gunkel1998list}  & Characteristics drawn from a set, e.g. choose one of Peaceful, Fearful, Erratic, Absentminded, Miserable, Skeptical.
\\ \hline

Character Tropes &
\cite{bamman2013learning} \cite{chu2018learning} & Descriptions such as Corrupt Corporate Executive, Retired Outlaw and Lovable Rogue.
\\ \hline

Descriptive Sentences & \cite{bogatu2015conversational}
\cite{zhang2018personalizing} & e.g. `I am an artist; I have four children; I recently got a cat; I enjoy walking for exercise; I love watching Game of Thrones.'
\\ \hline

Attribute-Value Pairs & \cite{joshi2017personalization} \cite{qian2018assigning} &
e.g. Gender: Male, Age: Young, Favourite Food Item: Fish and Chips.
\\ \hline \hline

\textbf{Implicit Personality Schemes} & \textbf{Definition} &  \textbf{Explanation} \\
\hline

Monologues from TV/Film/Other &
\cite{li2016persona} \cite{mazare2018training} & e.g. blog posts define personality.
\\ \hline

Dialogues from TV/Film/Other & \cite{li2016persona} \cite{chu2018learning} \cite{shawar2005chatbot} & e.g. TV show transcripts define personality. \\ \hline \hline

\textbf{Other Related Measures} & \textbf{Definition} & \textbf{Explanation} \\
\hline

Templates &
\cite{oraby2018controlling} & Sentence generation templates incorporate personality. \\ \hline

Genre Text & \cite{mathews2018semstyle} \cite{niu2018polite} & Genre labels such as `Romance' imply personality.
\\ \hline 

Other Linked Factors & \cite{walkerwalker1997improvising} \cite{shum2018elizashum2018eliza} & e.g. age, gender, language, speaking style, level of knowledge, areas of expertise.
\\ \hline

Historical Personas & \cite{heller2005freudbot} & Collected information about a famous person e.g. Chopin, Freud.
\\ \hline
\end{tabular}
\label{table_personality_method}
\end{table}

Eysenck \cite{eysenck1953structure} extended the work of Jung, maintaining that there were three \textbf{Personality Dimensions}, introversion-extraversion, neuroticism-stability and psychoticism. Together, he contended, they can portray a wealth of information about a person's personality. Eysenck \cite{eysenck1958short} presented a twelve-question questionnaire for measuring extraversion and introversion. Eysenck's approach, comprising a series of different dimensions, numerical scores along those dimensions and a means of determining those scores via a questionnaire, is the basis of much modern work on personality, and has had a great influence on the design of conversational agents.

Norman \cite{norman1963toward} and Norman and Goldberg \cite{norman1966raters} further extended the personality model of Eysenck, resulting in a set of five personality traits: agreeableness, conscientiousness, extraversion, neuroticism/emotional stability and openness to experience. These are known as the \textbf{Big-5 or OCEAN}, and are extensively referred to in the literature of personality and conversation.
In related work, R. Cattell \cite{cattell1967functional} developed an extension of intraversion-extraversion into a \textbf{16 Personality Factor questionnaire (16PF)}. The factors may be described as outoing, intelligent, emotionally stable, assertive, happy-go-lucky, conscientous, venturesome, tender-minded, suspicious, imaginative, astute, apprehensive, experimenting, group independent, controlled, tense. The 16PF has been the subject of considerable study and has been widely used since its invention. A further discussion including recent developments can be found in H. Cattell and Mead \cite{cattell2008sixteen}.

\textbf{Interpersonal Reactivity Index (IRI)} \cite{davis1980interpersonal} is concerned with the relationship between empathy and personality. It consists of four measures: Perspective Taking (being able to take the viewpoint of other people), Fantasy (being able to project onto the actions and emotions of characters in play, books and films), Empathetic Concern (tending to feel sympathetic towards other people), and Personal Distress (a feeling of anxiety in a social context).


\textbf{Traits} are personality-like characterstics of a person (or chatbot) drawn from a set. Some researchers devise their own lists, e.g. Peaceful, Fearful, Erratic, Absentminded, Miserable, Skeptical. Others use existing lists, one of the most interesting and extensive being that of Gunkel \cite{gunkel1998list}.

\begin{table}[t]
\centering
\caption{Training datasets containing personality data (ordered by date, then alphabetically within date). See Section \ref{personality_datasets} for more details.}
\begin{tabular}{|p{0.28\linewidth} | p{0.15\linewidth} | p{0.45\linewidth} |}
\hline 
\textbf{Name/Link/Ref} & \textbf{Source} & \textbf{Description} \\
\hline

List of Personality Traits\tablefootnote{\url{http://ideonomy.mit.edu/essays/traits.html}} \cite{gunkel1998list} & Gunkel & List of 638 primary personality traits, 234 positive and 292 negative. See Image-Chat \cite{shuster2018image,shuster2018engaging} \\ \hline

Film Dialogue Corpus\tablefootnote{\url{https://nlds.soe.ucsc.edu/fc}} \cite{walker2012annotated} & IMSDb & 862 film scripts, 7,400 film characters, 664,000 dialogue lines. Tagged with features allowing personality to be inferred.\\
\hline

CMU Movie Summary Dataset\tablefootnote{\url{http://www.cs.cmu.edu/~ark/personas/}} \cite{bamman2013learning} &
Wikipedia, Freebase &
501 characters associated with 72 tropes (and much other information).\\ \hline

Speaker Model \cite{li2016persona} &
Twitter &
24,725,711 3-turn sliding window sequences from 74,003 users, 92.24 per user. Private dataset. \\ \hline

Speaker-Addressee Model \cite{li2016persona} &
Friends \par Big Bang Theory &
69,565 dialogue turns for 13 characters in TV shows. Private dataset. \\ \hline

Personalized-Dialog\tablefootnote{\url{https://github.com/chaitjo/personalized-dialog}}\tablefootnote{\url{https://www.dropbox.com/s/4i9u4y24pt3paba/personalized-dialog-dataset.tar.gz}} \cite{joshi2017personalization} &
Synthesised &
Synthetic personalised dialogues for booking a restaurant. Uses KB. Personas defined by gender, age, favourite food. ~6,000 dialogues, ~20 turns. \\ \hline

Character Trope Description Dataset\tablefootnote{\url{https://pralav.github.io/emnlp_personas/}} \cite{chu2018learning} &
TVTropes\tablefootnote{\url{http://tvtropes.org}} &
Text descriptions of tropes, including typical characteristics, actions, personalities.
\\ \hline



Image-Chat\tablefootnote{\url{https://parl.ai/projects/image_chat/}} \cite{shuster2018image,shuster2018engaging} &
Crowdworkers &
215 style traits, 201,779 3-turn dialogues each based on an image. \\ \hline

IMDB Dialogue Snippet Dataset\tablefootnote{\url{https://pralav.github.io/emnlp_personas/}} \cite{chu2018learning} &
IMDB Quotes &
15,608 quotes linked to tropes in the Character Tropes Dataset (see above). \\ \hline

Persona-Chat\tablefootnote{\url{http://parl.ai/downloads/personachat/personachat.tgz}}\tablefootnote{\url{https://www.kaggle.com/datasets/atharvjairath/personachat}}\tablefootnote{\url{https://huggingface.co/datasets/AlekseyKorshuk/persona-chat}} \cite{zhang2018personalizing} &
Crowdworkers &
1,155 5-line random personas. 10,907 dialogues, 12-18 turns, 162,064 utterances. \\ \hline

ConvAI2\tablefootnote{\url{https://huggingface.co/datasets/conv_ai_2}} \cite{dinan2019second} & Crowdworkers & Persona-Chat plus 100 personas and 1,015 dialogues. \\ \hline

Microsoft Personality Chat\tablefootnote{\url{https://github.com/microsoft/botframework-cli/blob/main/packages/qnamaker/docs/chit-chat-dataset.md}} \cite{microsoft2019personality} & Microsoft &
100 scenarios, 5 personalities, 9 languages. \\ \hline

Movie Character Attributes (MovieChAtt) Dataset \tablefootnote{\url{https://github.com/Anna146/HiddenAttributeModels}} \cite{tigunova2019listening} &
Crowdworkers &
Active characters from 617 film scripts in Cornell Movie-Dialogs Corpus\tablefootnote{\url{http://www.cs.cornell.edu/ cristian/Cornell_Movie-Dialogs_Corpus.html}} are labelled with Profession, Gender, and Age. Also includes labels for some Persona-Chat personas, and for some Reddit users. \\ \hline

PersonalDialog\tablefootnote{\url{https://github.com/silverriver/PersonalDilaog}} \cite{zheng2019personalized} & Weibo &
Chinese dialogues with personality traits for each speaker (recognised by classifiers they trained), in attribute-value form. 20.83 million dialogues, 56.25 million utterances, 8.47M million speakers. \\ \hline

Persuasion-ForGood\tablefootnote{\url{https://gitlab.com/ucdavisnlp/persuasionforgood}} \cite{wang2019persuasion} & Crowdworkers & 1,017 conversations persuading a person to denote to charity. 1,017 conversations pursuading a person to denote to charity. Incorporates participants' Big-5 personality, Moral Foundations, Schwartz Portrait Value, and Decision Making Style. \\ \hline

FriendsPersona Dataset\tablefootnote{\url{https://github.com/emorynlp/personality-detection}} \cite{jiang2020automatic} & Friends & 711 dialogues from first four seasons of Friends TV show. Tagged for Big-5. \\ \hline

Personality Emotion Lines Dataset\tablefootnote{\url{https://github.com/preke/PELD}} \cite{wen2021automatically} & FriendsPersona & Dialogues from TV shows with both Emotion and Big-5 labels \\ \hline

IT-ConvAI2\tablefootnote{\url{https://github.com/CCIIPLab/Persona_Extend/}} \cite{liu2022improving} & ConvAI2 & 1,595 ConvAI2 conversations asking about personas. \\ \hline
   
\end{tabular}
\label{table_datasets}
\end{table}

\begin{table}[t]
\centering
\caption*{Table 2 continued.}
\begin{tabular}{|p{0.28\linewidth} | p{0.15\linewidth} | p{0.45\linewidth} |}
\hline 
\textbf{Name/Link/Ref} & \textbf{Source} & \textbf{Description} \\
\hline

Chinese Personality Emotion Lines Dataset\tablefootnote{\url{https://github.com/slptongji/GERP}} \cite{zhou2023gerp} & PELD & Chinese translation of Personality Emotion Lines Dataset. Includes both emotion adn Big-5 labels. \\ \hline

JPersonaChat\tablefootnote{\url{https://github.com/nttcslab/japanese-dialog-transformers}} \cite{sugiyama2023empirical} &
Crowdworkers &
Based on Persona-Chat approach but Japanese. 100 personas, 5 sentences each. 5,000 dialogues, 12-15 turns. \\ \hline

PersonageNLG\tablefootnote{\url{https://nlds.soe.ucsc.edu/stylistic-variation-nlg}}
\cite{ramirez2023controlling} &
PERSONAGE program &
Big-5 personality traits, 88,855 restaurant recommendations. \\ \hline    
\end{tabular}
\label{table_datasets2}
\end{table}

In general, Tropes are plot devices and archetypes relating mainly to TV shows and films. The term appears to emanate from the TV Tropes website\footnote{\url{https://tvtropes.org}}. Essentially, Tropes are witty and pithy noun phrases which are used to talk about popular media. \textbf{Character Tropes} are Tropes which specifically characterise personalities in films, TV etc. Examples include Retired Outlaw and Lovable Rogue.

\textbf{Descriptive Sentences} (our term) are sets of sentences which concisely define a personality. The most famous example is the Persona-Chat dataset \cite{zhang2018personalizing} (see Table \ref{table_datasets} and Section \ref{personality_datasets}) where a personality is defined as five sentences, e.g. `I am an artist; I have four children; I recently got a cat; I enjoy walking for exercise; I love watching Game
of Thrones.'

\textbf{Attribute-Value Pairs} (our term) are a set of attributes and corresponding values, which define a personality, e.g. \{Gender: Male, Age: Young, Favourite Food Item: Fish and Chips\}. As this example shows, Attribute-Value Pairs can be used to specify profile-type information which is getting rather far from the traditional concept of personality.

So far, all the schemes are explicit (as defined by Zheng \cite{zheng2019personalized}) in that there is a specific data structure or feature which defines a personality. Table \ref{table_personality_method} continues with two implicit schemes, where there is no such data structure. These are \textbf{Monologues from TV/Film/Other} and \textbf{Dialogues from TV/Film/Other}. In these cases, the personality is defined by the information which can be gleaned from a set of texts by a particular person, or a set of dialogues, all of which involve a particular person.

The final section of  the table lists some related measures which can sometimes be seen in papers. \textbf{Templates} are datastructures, usually for generating text (e.g. Oraby et al. \cite{oraby2018controlling} which can incorporate personality information. \textbf{Genre Text} is text of a particular style (e.g. `Romance') which can characterise personality. \textbf{Other Linked Factors} includes information which is related to personality, but is perhaps not personality itself, such as age, gender, language, speaking style, level of knowledge, or areas of expertise. Finally, we include \textbf{Historical Personas} as a category, because there are numerous chatbots which have been designed to impersonate famous characters such as Frédéric Chopin \cite{chopin2010intelliwise} or Sigmund Freud \cite{heller2005freudbot}. In such bots, the historical person themself is the specification of the personality.

\section{Personality Datasets} \label{personality_datasets}

In most cases, training of a chatbot with personality involves a suitable dataset including suitable personality data. As a result, a number of interesting datasets have been developed. These are summarised in Table \ref{table_datasets}. The datasets are ordered by date, and, within the same year, they are alphabetically ordered. Download URLs are provided in the table for each dataset, where these are publicly available.

Gunkel \cite{gunkel1998list} published a list of 638 primary personality traits, comprising 234 positive, 292 neutral, and 292 negative. Each trait is a single word. Interestingly, there seems to be no information about how these were arrived at, or what project they were intended for: Gunkel was a visionary figure, motivated purely by his interest in scholarly and academic matters. This list of personality words is one the longest in the literature. Examples of the 234 positive traits are Admirable, Brilliant, Charismatic, Dedicated, Eloquent, Forgiving, Gentle, and Honest. Examples of neutral ones are Absentminded, Boyish, Casual, Dreamy, Enigmatic, Frugal, Guileless, and High-spirited. Examples of the negative traits are Argumentative, Brutal, Calculating, Destructive, Egocentric, Foolish, Greedy, and Hostile. Naturally, there are many words for each letter of the alphabet in each of the three categories; these examples are included to give a flavour of the traits in Gunkel's list. These traits have been used in various projects, notably Image-Chat \cite{shuster2018image,shuster2018engaging} (see below).

Marilyn Walker at University of California Santa Cruz has played a major role in the development of personality models for natural language processing, and in the development of conversational agents embodying personality. The Film Dialogue Corpus \cite{walker2012annotated} comprises 862 transcripts of films from the Internet Movie Script Database\footnote{\url{http://www.imsdb.com/}}. Dialogues involve 7,400 different characters, 664,000 lines in total. The dialogues have then been tagged with various kinds of information. First, characters are labelled, based on properties of the films, namely Genre, Director, Year, and on the gender of the character. Second, the lines of each character are grouped together and analysed for various personality-indicating features, including Basic (number of tokens spoken by a character, number of turns),  Sentiment Polarity (on a scale of Positive to Negative), Dialogue Act, Passive Sentence Ratio, Linguistic Inquiry Word
Count (LIWC) word categories, and others. The Film Dialogue Corpus contains all this information. It does not specify personality as such; however, the authors show how to use this information to train a model which can infer the personality of each character. Finally, the trained model is used to control parameters of their PERSONAGE generator, which can generate versions of sentences with different personalities.

The CMU Movie Summary Dataset \cite{bamman2013learning} follows a similar approach and uses information from Wikipedia and Freebase. 42,305 film plot summaries are taken from the Wikipedia. These contain phrases such as `rebel leader Princess Leia'. the texts are analysed using CoreNLP\footnote{\url{http://nlp.stanford.edu/software/corenlp.shtml}} in order to extract Agent verbs, Patient verbs and Attributes. The authors argue that these are important in determining personality because they imply the actions carried out by a character, the actions carried out by others on the character, and the qualities or attributes of these actions.

The next step is to use Freebase\footnote{\url{http://download.freebase.com/datadumps/}} which contains detailed information about films and the characters within them. In particular, films are tagged with genre categories such as `Epic Western', and characters are linked to the actors who play them, with biographical information on these actors also being available.

According to the authors, a persona can be linked to three word distributions, concerning agent, patient and attribute. 
They use two models (Dirichlet Persona Model and Persona Regression Model) to find clusterings of words to topics (e.g. `strangle' is a type of Assault), topics to personas (e.g. VILLIANS perform many Assaults), and characters to personas (e.g. Darth Vader is a VILLAIN).

TV Tropes\footnote{\url{http://tvtropes.org}} contains common `Tropes' submitted by users in relation to TV, film and fictional works. Tropes are noun phrases which aptly characterise aspects of a narrative or character. The authors extract 72 tropes which can be associated with a character, phrases such as THE CORRUPT CORPORATE EXECUTIVE, or THE HARDBOILED DETECTIVE. These are then manually linked to 501 individual film characters, as identified earlier. This gold-standard data is then used to evaluate the two topic models.

The dataset itself consists of (1) Information for 81,741 films, obtained from Freebase, including name, release date and genre; (2) Information about 450,669 characters in these films, linked to the above, including film, character data and links to Freebase data; (3) 72 different character Tropes, together with 501 film characters, each assigned a Trope; (4) 970 character names in films, linked to 2,666 Trope instances.

Overall, this is an important dataset because it is the first which links Tropes to characters in a way which can be used for training a chatbot with personality. In addition, of course, it is introducing the idea of using Tropes to define personality alongside other approaches (see Tables \ref{table_personality_method} and \ref{table_personality_specification}).

The Speaker Model Dataset is discussed by Li et al. \cite{li2016persona}. It contains 24,725,711 3-turn sequences from 74,003 users, derived from Twitter. the Speaker-Addressee model contains 69,565 dialogue turns for 13 characters in TV show transcripts. These datasets appear to be private, but are included in the table for comparison. The use of these datasets is described below (Section \ref{chatbots_with_personality}).

Joshi et al. \cite{joshi2017personalization} proposed the The Personalized-Dialog dataset which is based on the well-known bAbI dialogue dataset \cite{bordes2016learning}. The original dataset uses a knowledge base of restaurants. For each there is a cuisine (10 types, e.g. English), location (10 e.g. Tokyo), a price range (cheap, moderate, expensive), rating (1-8), and party size (2, 4, 6, 8). Each restaurant also has an address and a phone number. The dataset consists of a set of synthetic structured dialogues, based on the knowledge base. The aim is to enquire about restaurants and book a table. Utterances in the dataset are created artificially using patterns, 43 for the user and 20 for the bot.

In their work, Joshi et al. produce a synthetic personalised dataset which is modelled on the original. First, five attributes are added to each restaurant in the KB, cuisine type (vegetarian, non-vegetarian), speciality dish, social media links, parking information, and public transport information. Second, each dialogue starts with a user profile, specifying gender (m/f), age (young, middle-aged, elderly), and favourite food. Thirdly, the linguistic patterns are augmented to take into account profile information; For each of the 15 bot patterns, there are now 6 variants (2 genders * 3 ages) with different usage for each. For example, for the older ages, a more formal style is adopted. The user patterns remain the same.

As in the original work \cite{bordes2016learning}, the dataset is created algorithmically, but the decisions now make reference to the profile information at the start of each dialogue. For example, when offering restaurants to the user, they are proposed based on their rating, with additional points for matching the user's diet preference (vegetarian or not) and favourite food. Thus, choice of proposed restaurants is now influenced by user profile. Similarly, depending on the age and gender of the user, the patterns used to create the bot's utterance are chosen accordingly. In addition, when asked for contact information, the bot returns social media if young, or phone if middle-aged/elderly. In a similar way, when asked for directions, the bot returns address and public transport information if the restaurant is cheap, or address and parking if it is moderately priced/expensive.

The dataset is used to evaluate various models, including rule-based systems, trained embeddings, and end-to-end memory networks.

In general, this work shows how a personalised restaurant dataset can be created algorithmically, as a first step towards producing training data which is tailored to a person's profile.

Chu et al. \cite{chu2018learning} make use of the CMU Movie Summary Dataset \cite{bamman2013learning} discussed above, as well as creating two new datasets. To build the IMDB Dialogue Snippet Dataset, they extract quotations for selected film characters from the IMDB Quotes website. There are 13,874 quotes for training, and 1,734 each for validation and testing. A quote may contain multiple dialogue turns; naturally, one of the characters in the quote must be the person of interest. Next, they create the Character Trope Description Dataset using information from TVTropes\footnote{\url{http://tvtropes.org}}. Along with each Trope there is a textual description. These two datasets are aligned, so that for a character we can see the quotes, the Trope of the character (e.g. Lovable Rogue) and the Trope description. Note that the aim of the work here is to characterise Personas in a pre-existing dialogue, not to produce a chatbot.


The starting point of the Image-Chat dataset \cite{shuster2018image} is 215 style traits which have been selected from a set of 638 traits created by Gunkel\footnote{http://ideonomy.mit.edu/essays/traits.html} (see Section \ref{personality_datasets}. The original 638 traits are already divided into three classes: positive, neutral, and negative, and selections are made from these to make up the set of 215. Traits which are applicable to captioning are chosen, for example sweet, happy (positive), old-fashioned, skeptical (neutral), anxious, childish (negative). Traits considered not applicable to captions are not used (e.g. allocentric, insouciant). The images used come from the YFCC100M Dataset.

The dataset is created using crowdworkers. In order to generate a dialogue, an image is first chosen at random. Two crowdworkers are chosen for the dialogue and each is assigned one style trait at random. They are then asked to create a three-turn dialogue based on the image. The emphasis is on the dialogue being interesting and engaging, and there is no requirement to refer to specific factual aspects of the image.

In a quality control process, a subset of the data was inspected by further crowdworkers, who found that 92.8\% of the time the dialogue clearly fitted the image, 83.1\% clearly fitted the styles assigned to the crowdworkers, and 80.5\% fitted both the image and the styles. So, as with many other datasets created using the crowd, the participants proved in the main to be imaginative and capable.

Persona Chat \cite{zhang2018personalizing} consists of a set of 10,907 dialogues created by pairs of Mechanical Turk workers. The starting point is a set of 1,155 persona specifications, which were created in the first phase of the project. A persona here consists of five profile sentences (each 15 words or less) describing a person. Turkers were asked to create such personas at random, by following an persona provided. In the second phase, the collected personas were revised by further Turkers, in order to refine the wording, and hence to avoid later problems of repetition. In the third phase, Turkers were divided randomly into pairs and each person was assigned a random persona from the collected set. They were then asked to engage in dialogue with each other, each person following the `personality' of their assigned persona. Each turn in the dialogue was 15 words maximum, and dialogues consisted of 12-18 turns in total (6-8 for each participant).

In summary, Persona Chat consists of general information-seeking chat. It seems that the Turkers used the personas very skilfully, resulting in some convincing dialogues. The specification of personas as five sentences seemed to have worked well, even though they are created from random combinations of personal characteristics. Naturally a persona in the sense used here is not the same as a personality; the personas are random combinations of personal facts and characteristics. However, this work and its predecessor, Image-Chat (see above) have been extremely influential. The dialogues are remarkably plausible and flow very naturally. These datasets have been widely used to evaluate other models, and other datasets based on the same idea have also been developed (see JPersonaChat below). As a method of specifying personality, this seems to be the dominant method being used today. It is much richer than a traditional specification such as Big-5, it can be used in many types of dialogue model, it can be integrated into the model in different ways (e.g. prefixing the content sentences in the dialogue model by the persona sentences -- see Table 3, line 3 -- or conditioning on a vector created with the persona sentences). Moreover, five sentences seems to provide enough information while at the same time being practical.

ConvAI2 \cite{dinan2019second} was created for the ConvAI2 evaluation campaign. It consists of Persona-Chat plus an additional 100 personas and 1,015 dialogues, used for evaluation.

The Personality Chat Dataset \cite{microsoft2019personality} consists of 100 chat scenarios, each for 5 personalities, in 9 languages. The personalities are Professional, Friendly, Witty, Caring, and Enthusiastic.

Tigunova et al. \cite{tigunova2019listening} present the Movie Character Attributes (MovieChAtt) Dataset. This is based on the Cornell Movie-Dialogs Corpus \cite{danescu2011chameleons}. Characters who have 20 lines or more are analysed. The Internet Movie Database(IMDb) was used to look up the gender and age of actors who played characters in the films, and hence to assign these attribute-value pairs to the film characters. Crowdworkers then assigned Professions to the characters. They did this by looking up the characters in Wikipedia articles about the films. In addition, working with the Persona-Chat dataset \cite{zhang2018personalizing} (see Table \ref{table_datasets}) personas were labelled with Profession, Gender, and Family Status. The result was 1,147 personas with a profession label, 1,316 with gender and 2,302 with family status. Thirdly, the authors used Reddit discussions in the `iama' and `askreddit' subforums, extracted from a public crawl\footnote{\url{https://files.pushshift.io/reddit/}}. Users were labelled with Profession, Gender, Age and Family status using pattern-matching methods. There are about 400 users labelled with each of these four attribute-value pairs. The authors present four Hidden Attribute Model classifiers and evaluate them on their datasets.

The PersonalDialog dataset \cite{zheng2019personalized} consists of a large set of dialogues extracted from logs of Weibo, a well-known Chinese social media site. Weibo itself provides no explicit personality information. Instead, the authors extract this information from the dialogues, using classifiers which they designed and trained. The personality information is then specified as a set of attribute-value pairs. The attributes include age, gender, language, speaking style, level of knowledge, areas of expertise, accent, location, interest. This dataset is large: There are 20.83 million dialogues over 8.47 million users, with 56.25 million utterances in total.

Wang et al. \cite{wang2019persuasion} develop an interesting dataset, Persuasion-ForGood, which includes dialogues where one participant persuades the other to donate to charity (see Section \ref{personality_datasets}. The underlying personality representation is a vector of 23 dimensions, including Big-5 \cite{goldberg1992development}, Moral Foundations \cite{graham2011mapping}, Schwartz Portrait Value \cite{cieciuch2012comparison}, and Decision-Making Style \cite{hamilton2016development}. A classifier combining LSTM and Recurrent CNN was developed to recognise persuasion strategy. However, a persuasive chatbot was not developed in the work reported.

Their goals are to link information about a person to the outcome of the persuasion process, and to investigate which persuasion strategies are most effective, given information about a person, including their personality.

This dataset uses multiple personality indicators, drawn from four different personality classification systems, to create the 23-dimension vector referred to above. This is an unusual approach, which we have not seen elsewhere.

IT-ConvAI2 \cite{liu2022improving} is based on ConvAI2 (itself based on Persona-Chat). It consists of 1,595 dialogues, selected from ConvAI2, which ask about persona information, with the persona information itself removed from the dialogue. Liu et al. \cite{liu2022improving} use this to evaluate their model.

JPersonaChat \cite{sugiyama2023empirical} is based on Persona-Chat cite{zhang2018personalizing} (see above) and replicates the approach for Japanese. There are 100 personas, specified with 5 sentences each, and 5,000 dialogues.

The PersonageNLG dataset \cite{ramirez2023controlling} contains 88,855 restaurant recommendations, which vary by Big-5 personality traits, namely agreeable, disagreeable, conscientious, unconscientious, and extrovert. Note, therefore, that this is text in different styles, it is not a dialogue dataset. The authors argue that Big-5 types can be seen in language at the lexical, syntactic and semantic levels \cite{mairesse2010towards}. The dataset is created automatically by the PERSONAGE program, which is described by Mairesse et al. \cite{mairesse2010towards}.

In Personage, there are both lexical and syntactic variations. It uses aggregation operations and pragmatic operations. Aggregation joins texts together, e.g. Extrovert person may use `with', `and', `also'. Pragmatic operations include addition of qualifying words such as `rather' before adjectives such as `expensive'.

The essence of this work is that it links Big-5 personality types to characteristics of language at the lexical, syntactic and semantic levels. It also describes the PERSONAGE program which is a statistical generator of text in different personalities. This is an unusual approach, as other researchers either specify personalities in advance (e.g. Persona-Chat) and then generate dialogues with those personalities using crowdworkers, or they capture natural dialogues from Blogs, social media sites, films or TV programs, and then retrospectively assign personality specifications to the users/characters.


\section{Chatbots with Personality} \label{chatbots_with_personality}



We now summarise some of the recent work, organising the analysis first around the personality model used (Table \ref{table_personality_specification} and, within that, by the method of embodying the personality in the chatbot (Table \ref{table_personality_embody}). See also Table \ref{table_personality_topics}. Please note that this discussion is therefore not chronological, so that important earlier work using one personality model may be discussed after later work using another personality model.

\begin{table}[t]
\centering
\caption{Personality specifications used in chatbot models.}
\begin{tabular}{|p{0.32\linewidth}|p{0.61\linewidth}|} 
\hline 
\textbf{Pre-existing Personality Schemes} & \textbf{Used By} \\
\hline

Introversion / Extraversion & \cite{shumanov2021making} \\ \hline

Personality Dimensions & \\ \hline

Big-5 / OCEAN & \cite{ball2000emotion}\cite{mairesse2011controlling} \cite{oraby2018neural} \cite{oraby2018controlling} \cite{zhou2019trusting} \cite{shumanov2021making} \cite{li2022prompt} \cite{barriere2022wassa} \cite{jiang2023personallm} \cite{safdari2023personality} \cite{zhou2023gerp} \\ \hline

16 Personality Factor (16PF) & \\ \hline

Interpersonal Reactivity Index (IRI) & \cite{li2022prompt} \\ \hline \hline

\textbf{Chatbot Personality Schemes} & \textbf{Used By} \\
\hline

Traits & \cite{shuster2018image} \cite{zhou2020design}
\\ \hline

Character Tropes &
\cite{bamman2013learning} \cite{chu2018learning}
\\ \hline

Descriptive Sentences & \cite{bogatu2015conversational} \cite{mazare2018training}
\cite{zhang2018personalizing} \cite{dinan2019second} \cite{song2019exploiting} \cite{wolf2019transfertransfo} \cite{lin2020xpersona} \cite{majumder2020like} \cite{cao2022model} \cite{firdaus2022enjoy} \cite{liu2022improving} \cite{xu2022diverse} \cite{sugiyama2023empirical}
\\ \hline

Attribute-Value Pairs & \cite{joshi2017personalization} \cite{yang2017personalized} \cite{qian2018assigning} \cite{luo2019learning} \cite{zheng2019personalized} 
\\ \hline \hline

\textbf{Implicit Personality Schemes} & \textbf{Used By} \\
\hline

Monologues from TV/Film/Other &
\cite{al2016conversational} \cite{li2016persona} \cite{mazare2018training}
\\ \hline

Dialogues from TV/Film/Other & \cite{shawar2005chatbot} \cite{shawar2005using} \cite{li2016persona} \cite{kottur2017exploring} \cite{nguyen2017neural} \cite{zhang2017neural} \cite{chu2018learning} \cite{madotto2019personalizing} \cite{tigunova2019listening} \cite{ma2021one} \cite{qian2021learning} \cite{zhong2022less} \\ \hline \hline

\textbf{Other Related Measures} & \textbf{Used By} \\
\hline

Templates &
\cite{oraby2018controlling} \cite{oraby2018neural} \\ \hline

Genre Text & \cite{mathews2018semstyle} \cite{niu2018polite}
\\ \hline 

Other Linked Factors & \cite{walkerwalker1997improvising} \cite{shum2018elizashum2018eliza}
\\ \hline

Historical Personas & \cite{heller2005freudbot} \cite{mehta2007developing} \cite{chopin2010intelliwise} \cite{weitz2014meet} \cite{bogatu2015conversational}
\\ \hline
\end{tabular}
\label{table_personality_specification}
\end{table}

\begin{table}[ht]
\centering
\caption{Methods to embody personality in chatbot models. The categories used are broad, and they are not exclusive, so a model can be cited on several lines. The table is indicative only, to show general trends.}
\begin{tabular}{|p{0.32\linewidth}|p{0.61\linewidth}|} 
\hline 
BERT & \cite{zhong2022less} \\ \hline

CNN & \cite{qian2021learning} \cite{zhong2022less} \\ \hline

Condition output on personality vector & \cite{luan2017multi} \cite{yang2017personalized} \\ \hline

Generative Adversarial Network & \cite{firdaus2022enjoy} \\ \hline

GPT & \cite{cao2022model} \cite{liu2022improving} \cite{xu2022diverse} \cite{jiang2023personallm} \\ \hline

Joint Training (parameter sharing) & \cite{luan2017multi} \\ \hline

Memory Network (incl. End-to-End) & \cite{joshi2017personalization} \cite{zhang2018personalizing} \cite{luo2019learning} \cite{song2019exploiting} \cite{ma2021one} \cite{firdaus2022enjoy} \\ \hline

Prefix with descriptive sentences & \cite{joshi2017personalization} \cite{shuster2018image} \cite{wolf2019transfertransfo} \cite{cao2022model} \\ \hline

Prompt & \cite{li2022prompt} \cite{jiang2023personallm} \cite{ramirez2023controlling} \cite{safdari2023personality} \\ \hline

Seq-to-Seq & \cite{luan2017multi} \cite{nguyen2017neural} \cite{yang2017personalized} \cite{zhang2017neural} \cite{oraby2018neural}\cite{oraby2018controlling} \cite{zhang2018personalizing} \cite{zheng2019personalized} \cite{ma2021one} \cite{zhou2023gerp} \\ \hline

Symbolic Template & \cite{heller2005freudbot} \cite{mehta2007developing} \cite{chopin2010intelliwise} \cite{oraby2018controlling} \cite{oraby2018neural} \cite{zhou2020design} \\ \hline

Transformer & \cite{shuster2018image} \cite{madotto2019personalizing} \cite{wolf2019transfertransfo} \cite{majumder2020like} \cite{ma2021one} \cite{qian2021learning} \cite{cao2022model} \cite{liu2022improving} \cite{zhong2022less} \\ \hline

Variational AutoEncoder & \cite{song2019exploiting} \\ \hline

\end{tabular}
\label{table_personality_embody}
\end{table}

\subsection{Introversion\/Extraversion}

The aim of Shumanov and Lester \cite{shumanov2021making} is to investigate whether congruent consumer-chatbot personality will benefit the user's engagement with the chatbot and improve the user's purchasing behaviour in a commercial setting. The work focuses on Big-5 personality, and specifically Introversion-Extraversion. Personality was manipulated using word-frequency usage data based on the Linguistic Inquiry and Word Count word association method \cite{tausczik2010psychological}. Personality recognition was carried out using an ML classifier. 57,000 chatbot dialogues with live users was carried out in an experimental setting. Interestingly, some were transcribed from voice dialogues. The domain was mobile phones, and the dialogues concerned enquiries about recharging credit, account balances and billing. The personality of users (just introversion-extraversion) was determined after the dialogue had taken place, with a classifier \cite{schwartz2013personality, plank2015personality, arnoux201725}. The method uses GloVe embeddings, combined for a text, which are input to a Gaussian Processes classifier. The chatbot personality was manipulated in terms of vocabulary usage to be either Introvert or Extravert. When a user contacted the chatbot to make an enquiry, they were assigned at random to either the Introvert version or the Extravert version. Purchasing behaviour was measured by the number of recharge services bought, while Engagement was measured by average duration of interaction with the chatbot Both purchasing and engagement improved when the personalities of the user and the chatbot were the same, according to their findings. 

\subsection{Big-5}

Xing and Fern{\'a}ndez \cite{xing2018automatic,xing2018examining} work with the seq-to-seq personality model of Li et al. \cite{li2016persona}. They use the Big-5 model of personality, pre-train on the OpenSubtitles dataset \cite{tiedemann2009news}, and then train on transcripts from the TV shows `Friends' and `The Big Bang Theory'. For each character, they estimate the Big-5 personality by applying the recogniser of Mairesse et al. \cite{mairesse2007using} to a sample of a character's utterances in the collection. This classifier gives a vector with scores for each Big-5 trait (see Table \ref{table_personality_method}). Now, the evaluation is interesting. Based on the outputs of their seq-to-seq model, they classify a sample of these into Big-5 vectors, again using the Mairesse tool. Finally, they compare these to the Gold-standard vectors; more specifically, they use a Support Vector Machine (SVM) classifier to try to establish the Gold-standard classes from those produced by the model. Results returned are above the baseline, suggesting the potential of this approach.

Wang et al. \cite{wang2019persuasion} develop an interesting dataset,  Persuasion-ForGood, which includes dialogues where one participant persuades the other to donate to charity (see Section \ref{personality_datasets}). The underlying personality representation is a vector of 23 dimensions, including Big-5 \cite{goldberg1992development}, Moral Foundations \cite{graham2011mapping}, Schwartz Portrait Value \cite{cieciuch2012comparison}, and Decision-Making Style \cite{hamilton2016development}. A classifier combining LSTM and Recurrent CNN was developed to recognise persuasion strategy. However, a persuasive chatbot was not developed in the work reported.

Li et al. \cite{li2022prompt} present a method for using prompts in order to identify personality. This was a participant at WASSA@ACL-2022 \cite{barriere2022wassa}. The personality methods are Big-5 and IRI (Table \ref{table_personality_method}). The prompt is a fixed template which is joined to the original input for learning. Data augmentation is used and an ensemble method is adopted, working over 3 pre-trained models.

The extensive work carried out by Marilyn Walker at University of California Santa Cruz has tended to focus on personality specified using Big-5 (see Table \ref{table_personality_method}).
The work of Oraby et al. \cite{oraby2018neural} combines symbolic and neural network approaches to personality in an interesting way. The starting point is the PERSONAGE corpus \cite{oraby2018controlling} which contains reports on restaurants expressed in different Big-5 personalities. They then train a seq-to-seq model, not on the texts but on the symbolic dialogue acts and associated parameters (e.g. which personality it is). At the end, a symbolic post-processing phase converts the dialogue act back into a sentence. In essence, when testing, if a combination of personalities is used, the output will combine qualities of those personalities.

Oraby et al. \cite{oraby2018controlling} compare three different seq-to-seq models which specify personality in different ways. The No-supervision model has simple input and output. The Token model encodes the personality as a single token. The Context model encodes the 36 style values, based on 5 aggregation operations and 19 pragmatic markers. In the results, the Context model is the best performing.

In Ramirez et al. \cite{ramirez2023controlling}, a prompt-based approach is used. The Big-5 types adopted are agreeable, disagreeable, conscientious, unconscientious, and extravert.
There are two types of prompt: (1) Data-to-text -- generating text from a representation which includes personality; (2) Textual-style-transfer -- converting the meaning representation to textual form. They train a BERT model on 4,000 texts from the PersonageNLG dataset, i.e. 800 per each of the 5 Big-5 personalities.
The Jurassic-1 Jumbo 175B parameter PLM is used for their experiments; this is an autoregressive language model. Training data-to-text gives better results.

In the GERP model, Zhou et al. \cite{zhou2023gerp} aim to produce a response, based on the personality, which is correct in emotion. Personality is based on the Big-5 scheme.

Jiang et al. \cite{jiang2023personallm} work with Large Language Models (LLMs) ChatGPT and GPT-4. They first create personalities based on Big-5 and use them to make the LLMs take the human 44-item personality test including writing a story. The test results produced by the LLMs are assessed for their Big-5 personalities, both using automatic and human methods. The results of the evaluation are consistent with the originally intended personalities. 

Safdari et al. \cite{safdari2023personality} present a method for testing the personality on LLMs. They use the IPIP-NEO [33] version of the Revised NEO Personality Inventory [19] as well as Big-5. Like Jiang et al. \cite{jiang2023personallm} They find that the LLMS can be made to mimic the specified personality under these tests.

\subsection{Interpersonal Reactivity Index (IRI)}

The work of Li et al. \cite{li2022prompt} is described above under Big-5. The same method, based on prompts, is used to predict IRI.

\subsection{Traits}

Shuster et al. \cite{shuster2018image} first present the Image-Chat dataset (Table \ref{table_datasets}) which is described in Section \ref{personality_datasets}. As we saw, it consists of a set of images, where for each there is a dialogue between two crowdworkers. Each crowdworker is assigned a Gunkel personalty trait \cite{gunkel1998list}. The generative dialogue model comprises three components, an Image Encoder, a Style Encoder, which transforms a trait into a distributed vector, 500 dimensions for retrieval, 300 for generation, and a Dialogue Encoder, based on Transformers \cite{vaswani2017attention}. The system is pre-trained on a large Reddit dialogue dataset, using the method of Mazare et al. \cite{mazare2018training}. The encoded image is added in the form of a token at the end of the Transformer encoder output. The trait encoding is added to the beginning of the dialogue history which is input to the Dialogue Encoder. This is therefore a form of `Prefix with descriptive sentences' in Table \ref{table_personality_embody}. Human evaluation is carried out, using images with associated dialogue which are not in the Image-Chat dataset.

The social assistant of Weitz \cite{weitz2014meet} and Zhou et al. \cite{zhou2020design} is outlined in the `Other' subsection (see below). However, the personality of this chatbot is also partly specified in terms of Traits (e.g. Figure 5 on page 63).

\subsection{Character Tropes}

Bamman et al. \cite{bamman2013learning} first used tropes to characterise personalities in film dialogues (see Section \ref{personality_datasets}). Chu et al.  also built a personality classifier based on tropes. However, neither built a dialogue system as such.

\subsection{Descriptive Sentences}


Zhang et al. \cite{zhang2018personalizing} first present the Persona-Chat dataset (see Table \ref{table_datasets}, Section \ref{personality_datasets}). They then discuss several possible models, some using ranking to select and output, others using a Seq-to-Seq model to generate the output. In particular, one model uses Seq-to-seq augmented with a memory network in which the five profile sentences from Persona-chat are individual entries. The models are evaluated by predicting the next utterance, and there is also a human evaluation based on Fluency, Engagingness, Consistency and Persona Detection. On Persona Detection, a model using ranking along with Profile Memory appears to perform the best, followed by the Seq-to-seq with Profile Memory.

Mazar{\'e} et al. \cite{mazare2018training} build on the ideas of Persona-Chat \cite{zhang2018personalizing}, which uses 1,155 5-line profiles created by crowdworkers (see Section \ref{personality_datasets}). The general aim here is to predict responses in a conversation, based on the persona of the speaker. A far bigger set of random personas is created as follows. A Persona is constructed using a REDDIT collection of 1.7 billion comments. All comments from a particular user are grouped together. Next, those between 4 and 20 words, containing 'I' or 'my', at least one verb and at least one noun/pronoun/adjective are chosen. A persona is limited to {\em N} sentences chosen from the resulting set. The main method for doing this is by using a classifier based on a bag of words, which is trained to select Persona-Chat sentences and to reject random comments. There are therefore 4.6 million personas.

A one-hop memory network is used, where the context sentences are the query, and the persona sentences are the memory. The representation, which is a combination of the context sentence and the persona, is compared to a set of candidate responses, and the one which matches best is chosen as the response. Matching is performed using the dot product. When testing, a Transformer-based model incorporating personas works better than the baselines. In general, this work shows that training with a `Descriptive Sentences' approach (Table \ref{table_personality_specification}) can be taken to extremes, and still give good results when evaluated on the original Persona-Chat data. 

The aim of Song et al. \cite{song2019exploiting} is to create more diverse responses from a CA, while incorporating personality. This work is done using the Persona-Chat dataset with the additional test set from the ConvAI2 evaluation. Personality is thus expressed as five sentences. The model used is based on Conditional
Variational AutoEncoders [Zhao2017] and uses aspects of MemN2N [Sukhbaatar 2015)]to incorporate an external memory for the persona information.

Wolf et al. \cite{wolf2019transfertransfo} describe TransferTransfo, a participant at the Conversational Intelligence Challenge 2 (ConvAI2)\footnote{\url{http://convai.io/}}. The model is based on the Generative Pre-Trained Transformer \cite{radford2018improving}. Following pre-training on BooksCorpus \cite{zhu2015aligning}, the model is trained using Persona-Chat \cite{zhang2018personalizing}. The persona is specified by prefixing the dialogue sentences with the corresponding descriptive sentences from Persona-Chat.

The key to the work of Cao et al. \cite{cao2022model} is to carry out various processes on previous dialogue data in order to simplify and improve its processing. These are (1) Data distillation to make data only contain useful sentences, (2) Data diversification to improve diversity of later responses, (3) Data curriculum \cite{bengio2009curriculum}, i.e. the use of the simplified data for initial training, followed by the original data. The data used is Persona-Chat, so this approach is based on `Descriptive Sentences' in our terminology. The approach can be used with different base models. Here they present versions using Transformers \cite{vaswani2017attention} and GPT2 \cite{cao2020pretrained}. Human evaluation is relative to Fluency, Coherence with dialogue history, and Persona consistency. With Transformers, results are better than the baselines.

The aim of Firdaus et al. \cite{firdaus2022enjoy} is to generate responses which match the persona and emotion of the user, working in the `Descriptive Sentences' paradigm. The model consists of a generator network, a memory network in which the persona information is encoded, and three discriminators, Empathetic, Persona, and Semantic. The generator is a hierarchical encoder-decoder using BiGRU. The decoder is a single-direction GRU. The three discriminators use GRU to find the distance between a generated response and the ground-truth, in respect of semantics, persona, and empathy. Experiments are conducted on ConvAI2 (based on Persona-Chat). Human evaluation is on the basis of Fluency, Relevance, Persona Consistency, and Emotion Appropriateness.

The key idea of Liu et al. \cite{liu2022improving} is to develop a way to retrieve a persona from a large collection which is consistent with an existing 5-line persona of the Persona-Chat form. In this way, the source material for a retrieval model can be enriched. For example, a query about a family situation cannot readily be answered if the existing persona does not specify this. However, the persona expansion must not accidentally contradict the existing persona. The model contains two parts -- (1) The Persona Ranking Model (PRM) ranks personas from the entire ConvAI set using natural language inference to prevent conflicts; (2) PS-Transformer selects the most suitable personas from these. PRM uses an existing pretrained natural language inference module \cite{gao2021adapting}. PS-Transformer is a Transformer model based on OpenAI GPT \cite{radford2018improving}. They propose a special dataset for evaluation: IT-ConvAI2 (see Section \ref{personality_datasets}). The model achieves good results on the ConvAI2 and IT-ConvAI2 datasets.

The intent of Xu et al. \cite{xu2022diverse} is to combine information from the persona (in a `Descriptive Sentences' paradigm) and from the dialogue history, in order to generate good responses. They use a Persona Information Memory Selection Network (PMSN) to do this. PMSN is an MLP, trained on previous dialogue data. GPT-2 is used within the generator. There is a multi-hop attention mechanism for previous dialogue information and for persona information. The model is trained on ConvAI2 data. The model can generate a more personalised response than the baselines, according to their evaluation.

\subsection{Attribute-Value Pairs}

Yang et al. \cite{yang2017personalized} perform personalisation based on user profiles. A seq-to-seq model is used. They use a form of transfer learning: Train first with a large general training dataset, then train with a smaller personalised dataset. They use Sina Weibo conversation data (with no user information), i.e. 4,435,959 post-response pairs (in 219,905 posts) crawled from Sina Weibo by Shang et al. \cite{shang2015neural}. They then crawled Sina Weibo themselves for tweets from users who had profile information, extracting the profiles as well. This led to 51,921 post-response pairs with profile data. The profile consists of User identity, Age (in five age bands), Gender (female, male or None), Education, and Location. This is placed in a one-hot vector and then converted to a distributed representation during training. The output of the seq-to-seq is conditioned on this distributed vector. In human evaluation the proposed personalised model outperformed the baselines.

Qian et al. \cite{qian2018assigning} present an approach based on an attribute-value approach to personality, which they define as `the character that the bot plays or performs during conversational interactions'. A key problem they want to address is inconsistent responses to the same question within a session, a well-known problem with chatbots. For example, in their Table 1: User: `Are you a boy or a girl?', Chatbot: `I am a boy.', User: `Are you a girl?', Chatbot: `Yes, I am a girl.'. This problem is addressed using profiles. The model is first trained on Chinese social media data. Next, profile information is created, consisting of key-value pairs for different personalities. When running a dialogue, the required profile is specified. If the chatbot is asked a question relating to one of the keys (e.g. 'how old are you?'), the value from the profile is substituted into the answer returned by the chatbot using the position detector. If the chatbot is asked a question not relating to the keys (e.g. 'how are you?) then the chatbot response is used with no substitution.

Luo et al. \cite{luo2019learning} are aiming at personalisation in a goal-oriented setting (specifically, restaurant reservation). They use an End-to-End Memory Network (MEMN2N [Bordes, Boureau, and Weston (2017]. Utterances in the dialogue are added to the memory as the dialogue proceeds. A user Profile is originally in attribute-value form, e.g. Gender: Male, Age: Young etc. In their Profile Model, a distributed version of the original profile is used. They also incorporate dialogue from similar users as a global memory. The Preference Model can learn a users preferences in respect of entities in the knowledge base. Combining these two models gives the best performance.

\subsection{Implicit -- Using Dialogues/Monologues}

Abushawar and Atwell \cite{shawar2005chatbot,shawar2005using} is a very early project in which the aim was to train a chatbot on corpus data. They used parts of the British National Corpus (BNC) to train ALICE chatbots in different domains such as Sport, World affairs, travel, media and food. They also used BNC London teenager and "loudmouth" transcripts. In such a way, their chatbots could take on different personalities and adopt different styles of language. The personalities are of course implicit, as there is no underlying personality model such as Big-5 etc.

Li et al. \cite{li2016persona} define a persona as a combination of background facts relating to a user, together with their language behaviour and style of interaction. Moreover, they state that the persona of a chatbot should adapt dynamically, depending on who the chatbot is interacting with. In this work, therefore, there is no explicit personality model (e.g. Big-5 or Character Tropes, say). Instead the personality is implicit, deriving from the collective behaviour of the same speaker over many dialogues.

They have two persona models, a single-speaker Speaker model and a dyadic Speaker-Addressee model. The Speaker model captures how the speaker behaves in general, i.e. their persona. The Speaker-Addressee model captures how the speaker behaves when talking to a particular person. This allows for the possibility that a person speakers differently to different people.

The dialogue model is seq2seq. It is trained by inputting the speaker ID and addressee ID along with the context utterance to get the response utterance. They also use domain adaptation training. First, train on a big conversational dataset with no personalisation. Then, train on a smaller dataset with personalisation.

The Speaker model is a vector representation which forms part of the target LSTM in the seq-to-seq model. The vector is learned during training, and the vector for a particular speaker is shared across all training instances for that speaker. Thus, the vector comes to learn the characteristics of a person, and can then condition the output produced for that person. In the Speaker-Addressee model, the principle is similar, but the vector encodes information about both the Speaker and the Addressee. Now, the vector is shared across all training instances between a particular speaker and a particular addressee.

The Speaker model is based on training using specific-speaker data extracted from Twitter. Speakers were selected who participated in between 60 and 300 3-turn conversations on Twitter within a six-month period. The resulting training set contained 24,725,711 3-turn sliding window sequences (i.e. context-message-response) for 74,003 users, having on average 92.24 turns in the dataset. In addition there are 12,000 3-turn conversations for development, validation and testing, 4,000 each. The Speaker-Addressee model is based on scripts from two American television series, {\em Friends} and {\em The Big Bang Theory}. Thirteen main characters were chosen from these series and used to collect a dataset of 69,565 turns. Development and test sets contain about 2,000 turns each, the remainder are for training. In this model, account is taken both of the speaker and the person to whom they are speaking, in developing the implicit persona of the speaker. Such a model can account for the persona of a person being different, depending on who they are speaking to.

This was one of the earliest attempts to introduce personas into a chatbot, and the results are somewhat inconclusive. The datasets are not public, but some information is included in Table \ref{table_datasets} for comparison.

The aim of Nguyen et al. \cite{nguyen2017neural} is to build chatbots to imitate characters in TV programmes -- Barney from How I Met Your Mother, Sheldon from The Big Bang Theory, and Michael from The Office and Joey from Friends. This is done using seq-to-seq model for each character and training it on different data. Because the text of these TV shows is not long enough (about 50,000 utterance-respose pairs), a very interesting form of transfer learning is used, which incorporate the Cornell Movie-Dialogs Corpus \cite{danescu2011chameleons} (about 200,000 pairs).
The training regime is as follows: Phase 1: 10k iterations on the Cornell corpus and all four TV shows; Phase 2: 5k iterations on just TV shows; Phase 3: 2k iterations with just utterances by a chosen TV character. Repeat this separately for each character, to create a different chatbot for each one. According to human evaluation, the personalised chatbots performed well.

Luan et al. \cite{luan2017multi} adopt joint training in order to achieve personalisation in a conversation agent. There are two models, a seq-to-seq model and an autoencoder. In this formulation, the autoencoder is also a seq-to-seq model. The first model is learning, in the usual way, to respond in a conversation while the autoencoder is trained to predict the input sequence itself. The two are trained with shared parameters so that both learn together and there is generalisation across the two. The seq-to-seq is trained on triples from Twitter. Single posts on Twitter (not interacting with others) of the top 20 users were considered as non-conversational data and used to train the autoencoder. The application domain was technical support. There were two variants of the task; both performed better than the baseline in responding to technical support requests, according to human evaluators.

Zhang et al. \cite{zhang2017neural} develop five different personalised seq-to-seq models. This is achieved in two stages (i.e. transfer learning). First, train all five chatbots on 1,000,000 post-response pairs from Chinese online forums. Then, train each chatbot on data specific to one user. To create this, five volunteers made available 2,000 messages from their chat history. These are responses with no posts. Therefore, they searched the main data for the closest response and then used the corresponding post to align with the personalised response. The result is five different chatbots, one per user. These were then evaluated by several methods. One is an interesting form of modified Wizard-of-Oz: First, the researcher and the chatbot both receive a tester's tweets. Both give a response, and the researcher decides which to return to the sender. Then the researcher asks the sender whether all replies were from the researcher (they don't know about chatbot). An Imitation Rate is calculated in this way.

The key idea of Madotto et al. \cite{madotto2019personalizing} is to use few-shot learning -- from a few previous dialogues with a particular user, predict their profile. Then they use this to adapt the CA to converse appropriately with that user. Essentially, they sample a batch of personas from the total set. For these, they partition conversation data for these personas into train and validation sets. Then train on them for some iterations. Then sample another batch of personas, and so on. Note that we never train on the personas themselves, only on the past conversations by these personas. They use Persona-Chat dataset. Each persona description has 8.3 unique dialogues. The dialogue model, they say, is a standard Transformer approach \cite{vaswani2017attention}. Results of their training method are comparable to those using the personas directly, and they claim that their dialogues are more consistent.

Qian et al. \cite{qian2021learning} and Ma et al. \cite{ma2021one} (see below) are in the same group and propose similar (but not identical) personalisation models. Qian et al. take a retrieval based approach, which selects the best response from a set of candidates. First, they find the language style of the user from their previous dialogues (social media posts and responses). Second, they match utterances from previous dialogues to the current dialogue, to find those that are related by topic. This influences the choice of response candidates. Finally, two forms of matching are combined to choose a response, matching with the personalised style and matching with the dialogue history. The essence of their architectures is an Attentive Module \cite{zhou2018multi} with single head attention which is based on Transformers \cite{vaswani2017attention}. It matches a query and key-value pair to an output. Unusually, the model uses CNN to extract matching features relating to a response candidate and historical responses. A multi-hop approach to retrieval is adopted to alleviated problems of ambiguity. In the final step, an MLP is used to combine the results of the two matches (style and previous dialogues). Evaluation is relative to Weibo and Reddit (see also Ma et al. below) and is claimed to be better than baselines.

Ma et al. \cite{ma2021one} assert that an approach based on explicit sentences (e.g. Persona-Chat) is not practical in large contexts. Instead, the follow the approach of implicit personality and learn profile information from previous dialogue data for a user. First, using previous interactions, they find the interests of a user, their interests and their writing style. A Transformer language model is used to construct this general user profile. Next, they build a post encoder for each user, which uses the user profile. This uses an RNN based on BiGRU. Third, a key-value memory network holds previous utterance-response pairs for the user. The network is used to create a dynamic profile which can attend to posts from the history which are relevant to the current dialogue. For decoding, there are two mechanisms. A word can be predicted from the general vocabulary or from the personalised vocabulary. To do this ideas from CopyNet \cite{gu2016incorporating} are used. For experiments, data from Weibo and Reddit is used. For evaluation, they use Bleu-1/2, Rouge-L, Dist-1/2 and human evaluation. With the evaluation by humans, the proposed model is judged superior to baselines on Readability, Informativeness and Personalisation.

Zhong et al. \cite{zhong2022less} present a personality model using implicit data extracted from dialogue history. Their Modeling and Selecting user Personality (MSP) model tries to extract the most useful persona information from a user's dialogue history. There are three key refiners: (1) The user refiner selects other users with similar interests to the current user. (2) The topic refiner searches for related topics in both the current user's history and those of the related users selected by (1). (3) The token refiner finds the best user profiles at the token level. Then, a generator produces responses, using user profiles and the input query. The user refiner uses a BERT model, then compares candidates with the dot-product. The topic refiner also uses a similar BERT approach, while the token refiner adopts Transformers. A Transformer decoder is used. There is also a sentence-matching component using CNN and RNN, which is used to train the token refiner. Evaluation is on Weibo and Reddit datasets.

\begin{table}[t]
\centering
\caption{Personality Research Topics.}
\begin{tabular}{|p{0.35\linewidth} | p{0.55\linewidth} |}
\hline 
\textbf{Topic} & \textbf{Discussed By} \\
\hline
Dialogues with personality & \cite{heller2005freudbot}  \cite{shawar2005chatbot}\cite{shawar2005using} \cite{mehta2007developing} \cite{chopin2010intelliwise} \cite{weitz2014meet}\cite{bogatu2015conversational} \cite{dodge2015evaluating}\cite{knight2016amazon} \cite{li2016persona} \cite{zhang2016combining} \cite{luan2017multi} \cite{nguyen2017neural} \cite{shevat2017designing} \cite{stinson2017surprising} \cite{yang2017personalized} \cite{zhang2017neural} \cite{chu2018learning} \cite{goode2018google} \cite{kang2018study} \cite{mazare2018training} \cite{oraby2018neural} \cite{oraby2018controlling} \cite{qian2018assigning} \cite{xing2018automatic} \cite{xing2018examining} \cite{zhang2018personalizing}\cite{luo2019learning} \cite{madotto2019personalizing} \cite{song2019exploiting} \cite{tigunova2019listening} \cite{wang2019persuasion} \cite{wolf2019transfertransfo} \cite{zheng2019personalized} \cite{majumder2020like} \cite{zhou2020design} \cite{ma2021one} \cite{qian2021learning} \cite{shumanov2021making} \cite{cao2022model} \cite{firdaus2022enjoy} \cite{li2022prompt} \cite{liu2022improving} \cite{xu2022diverse} \cite{zhong2022less} \cite{jiang2023personallm} \cite{ramirez2023controlling} \cite{safdari2023personality} \cite{zhou2023gerp} \\ 
\hline
Generating text in different styles & \cite{hovy1987generating}
\cite{walkerwalker1997improvising} \cite{mairesse2010towards} \cite{mairesse2011controlling} \cite{stinson2017surprising} \cite{zhang2017neural} \cite{zhou2017emotional} \cite{goode2018google} \cite{kang2018study} \cite{reed2018can} \cite{niu2018polite} \cite{oraby2018controlling} \cite{oraby2018neural} \cite{elsholz2019exploring} \cite{harrison2019maximizing} \cite{reif2021recipe} 
\\ \hline
Evaluation of personality & \cite{mairesse2011controlling} \cite{kuligowska2015commercial} \cite{liu2015two} \cite{knight2016amazon} \cite{stinson2017surprising} \cite{chinnakotla2018lessons} \cite{kang2018study} \cite{xing2018automatic} \cite{xing2018examining} \cite{sugiyama2023empirical}
\\ \hline
Personality identification &
\cite{mairesse2007using} \cite{tausczik2010psychological} \cite{yarkoni2010personality}\cite{walker2012annotated}  \cite{adali2012predicting} \cite{bamman2013learning} \cite{kosinski2013private} \cite{mahmud2013recommending} \cite{gou2014knowme} \cite{kern2014online} \cite{preotiuc2016studying} \cite{zhang2017neural} \cite{chu2018learning} \cite{xing2018automatic} \cite{xing2018examining} \cite{junior2019first} \cite{tigunova2019listening} \cite{zhou2019trusting} \cite{vora2020personality} \cite{beck2022mega}\cite{barnes2022proceedings}  \cite{barriere2022wassa} \cite{ramirez2023controlling}
\\ \hline
Coherent personality & \cite{yu2016strategy} \cite{zhang2016combining} \cite{qian2018assigning} \cite{xing2018automatic} \cite{xing2018examining}
\\ \hline
Personality in image captions & (\cite{mostafazadeh2017image}) \cite{shuster2018image} \cite{shuster2018engaging} \cite{huber2018emotional}
\\ \hline
Personality and user trust & \cite{neff2016talking} \cite{agrawal2018trustworthy} \cite{zhou2019trusting}
\\ \hline
Prompt methods & \cite{reif2021recipe} \cite{bach2022promptsource} \cite{li2022prompt} \cite{liu2023pre} \cite{ramirez2023controlling}
\\ \hline
Chatbot personality discussions & \cite{lessio2020toward} \cite{pradhan2021hey}
\\ \hline
\end{tabular}
\label{table_personality_topics}
\end{table}

\begin{table}[t]
\centering
\caption{Related Topics.}
\begin{tabular}{|p{0.35\linewidth} | p{0.55\linewidth} |}
\hline 
\textbf{Topic} & \textbf{Discussed By} \\
\hline
Early end-to-end chatbots & \cite{jafarpour2010filter} \cite{ritter2011data} \cite{banchs2012iris} \cite{shang2015neural}
\cite{sordoni2015neural} \cite{vinyals2015neural}
\\ \hline
Origins of neural question answering & \cite{andreas2016learning}
\\ \hline
General evaluation of chatbots & \cite{walker1997paradise} \cite{shawar2007chatbots} \cite{shawar2007different} \cite{kuligowska2015commercial} \cite{abushawar2016usefulness} \cite{liu2016not} \cite{lowe2016evaluation}  \cite{masche2018review} \cite{kvale2019improving} \cite{sinha2020learning} \cite{borsci2022chatbot} \cite{shumanov2021making} \cite{sugiyama2023empirical}
\\ \hline
Challenges and competitions & \cite{dinan2019second} \cite{barnes2022proceedings} \cite{barriere2022wassa}
\\ \hline
Conversational search & \cite{belkin1995cases} \cite{radlinski2017theoretical} \cite{joho2018cair}
\\ \hline
Search personalisation & \cite{matthijs2011personalizing} \cite{bennett2012modeling} \cite{dodge2015evaluating}
\\ \hline
Personality chatbot reviews &
\cite{ait2023power} \cite{ferreira2023review}
\\ \hline
General chatbot reviews & \cite{zue2000conversational} \cite{io2017chatbots} \cite{li2017conversational} \cite{zhang2016combining} \cite{gao2018neural}  \cite{masche2018review} \cite{shum2018elizashum2018eliza} 
\cite{csaky2019deep} \cite{gao2019neural} \cite{hussain2019survey} \cite{mnasri2019recent} \cite{poria2019emotion} \cite{pamungkas2019emotionally} \cite{santhanam2019survey} \cite{fernandes2020survey} \cite{roller2020recipes} \cite{rapp2021human} \cite{luo2022critical} \cite{singh2022survey} \cite{yan2022deep} \cite{scotti2023primer} \cite{xi2023rise} 
\\ \hline
Other related reviews & \cite{junior2019first} \cite{vora2020personality} \cite{beck2022mega} \cite{maithri2022automated} \cite{liu2023pre}
\\ \hline
Emotion and empathy in dialogue & \cite{byrnes2016bot} \cite{banchs2017construction} \cite{ghosh2017affect} \cite{gupta2017sentiment} \cite{zhou2017emotional} \cite{liu2018should} \cite{niu2018polite} \cite{rashkin2018towards} \cite{swayne2018want} \cite{huang2018automatic} \cite{ghosal2019dialoguegcn} \cite{huang2019emotionx} \cite{lin2019caire} \cite{lin2019moel} \cite{shin2019generating} \cite{naous2020empathy} \cite{zhou2020design}
\\ \hline
\end{tabular}
\label{table_related_topics}
\end{table}

\subsection{Other}

Several bots have been created in order to impersonate famous historical persons.
Heller et al. \cite{heller2005freudbot} discuss a chatbot developed in order to improve student learning. It is based on traditional pattern recognition using AIML templates. The bot imitates the personality of Sigmund Freud and is coded with various materials about him.

Mehta and Corradini \cite{mehta2007developing} present a chatbot which can converse about Hans Christian Andersen -- his life and the fairy tales he wrote. The bot is embodied in a 3D graphical character which can speak and make gestures. The purpose is to educate in an entertaining way. The implementation uses a dialogue-act detector to process the input, determine the intent (e.g. yes/no question) and respond accordingly. It is influenced by ELIZA \cite{weizenbaum1966eliza} and Alice \cite{abushawar2015alice}.

InteliWISE \cite{chopin2010intelliwise} developed a chatbot based on the composer Frederic Chopin. It could talk about Chopin, his life, adventures, compositions, journeys he made, people he was associated with, and so on. Details are scarce, but we can assume it was based on templates and pattern matching.

Weitz \cite{weitz2014meet} and Zhou et al. \cite{zhou2020design} describe a social assistant which is available through Weibo, the Chinese social media site. It is suggested that it has a distinct personality, can empathise with the user and that it has a sense of humour. The bot is able to respond to the same question on different occasions with a different tone. As of 2014, it had been used for 0.5 billion conversations, up to 200,000 of them being simultaneous. Implementation is based on a dialogue manager.

Bogatu et al. \cite{bogatu2015conversational} create an agent which can answer questions about historical figures like Albert Einstein and John Lennon. They use two methods: (1) A knowledge base created via an ontology, and (2) Answer selection. The agent is built using ChatScript. They create rules for it by extracting information from Wikipedia and DBpedia. There are two methods for doing this: First, extracting question-answer pairs from Wikipedia, matching the input with the question and outputting the answer; second, matching the question with Wikipedia text directly, converting the best sentence to first person and outputting it. Essentially, this work is stating Wikipedia facts about famous persons, using the first person. There is no formal model of personality, but it is implicit in the material extracted. It can be viewed as a specification via descriptive sentences, but is obviously quite different from Zhang et al. \cite{zhang2018personalizing} where sentences are written by crowdworkers.




\section{Relevant Reviews} \label{reviews}
In recent years there have been an increasing number of reviews relevant to chatbots and conversational agents (see Table \ref{table_related_topics}. We summarise them here.

Ait et al. \cite{ait2023power} have just published a review of personality-adaptive chatbots. Ferreira and Barbosa \cite{ferreira2023review} is a survey concerned with the effects on users of personality in chatbots.

Concerning the closely-related topic of emotion in conversation agents, Zhang et al. \cite{zhang2016combining} present a review of chatbots with emotion which touches on personality. Poria et al. \cite{poria2019emotion} present a detailed discussion of relevant work, focusing on the detection of conversational emotion. Also of interest, but further from the topic of this article, Maithri et al \cite{maithri2022automated} review methods for recognising emotion in EEG, facial information and speech systems.

Junior et al. \cite{junior2019first} is a survey of work which aims to extract personality traits from images, image sequences and video. Vora et al. \cite{vora2020personality} reviews work on personality prediction from social media sources.
Yan et al. \cite{yan2022deep} is an extensive recent review on dialogue systems which touches on personas etc. on p40-41.
Jin et al \cite{jin2022deep} review research on text style transfer (TST), going back to the 1980s. They mention the relevance of TST to the creation of chatbot personas, as outlined by McDonald and Pustejovsky \cite{mcdonald1985computational} and Hovy \cite{hovy1987generating}, and refer briefly to relevant work such as Li et al. \cite{li2016persona}, Zhang et al. \cite{zhang2018personalizing} and Shuster et al \cite{shuster2018image} [they reference as 2020 ACL] (see our discussion below).

Concerning general reviews on Conversation Agents, one of the earliest, dealing with the pre-neural era, is Shawar et al. \cite{shawar2007chatbots}. Deshpande et al. \cite{deshpande2017survey} and Hussain et al. \cite{hussain2019survey} are general surveys. Fernandes et al. \cite{fernandes2020survey} is a short survey focusing on practical chatbots in the pre-neural paradigm. Csaki, in his dissertation at Budapest University of Technology and Economics \cite{csaky2019deep} presents a detailed and clear review of previous work on neural chatbots work as well as explaining the underlying theory. Ni et al. \cite{ni2023recent} is a comprehensive and detailed survey of theory and practice in neural network dialogue systems. Rapp et al. \cite{rapp2021human} is a general review of how users interact with chatbots. Singh and Beniwal \cite{singh2022survey} is another recent review. 

Xi et al. \cite{xi2023rise} is a review of conversation agents based on LLMs. Finally, Liu et al. \cite{liu2023pre} is a survey on prompting methods in general which are used for an increasing number of NLP tasks, including personality detection \cite{li2022prompt}.

\section{Conclusions}

CAs and personality has become a very active area of research. What can we conclude about what has been done so far, and what will happen next?

Our first observation is that the development of the Image-Chat \cite{shuster2018image} and Persona-Chat \cite{zhang2018personalizing} datasets has been one of the most influential and remarkable areas of work. Underlying this is the idea of using Descriptive Sentences to define personality (Tables \ref{table_personality_method}, \ref{table_personality_specification}). The transparency of textual information to express personality has great advantages as well as future potential. On the other hand, implicit approaches based on the analysis of previous dialogues etc. are becoming increasingly sophisticated (e.g. \cite{ma2021one}) and can perform in an unsupervised way on large datasets.

Secondly, there has been huge progress in embodying CAs with personality within a short period of time. Many very ingenious and imaginative schemes have been developed and many remarkable datasets have been created.

Thirdly, we can see some hints concerning NN architectures by glancing at Table \ref{table_personality_embody}. Seq-to-Seq and Transformer methods underly almost all work, and various kinds of Memory Network are also increasingly used.

However, many of the actual results are somewhat inconclusive. This is partly linked to the difficulty of assessing -- especially automatically -- the consistency and effectiveness of the personality. A lot of automatic evaluation falls back on simple comparison with a baseline based on, for example, the gold standard next utterance. Regarding human evaluation, it is expensive, and, as yet, it is not quite clear exactly what we are measuring when it comes to personality.

So, turning to the future, we can predict that many new and remarkable datasets will appear, based on the personality schemes discussed in this review. Both explicit and implicit personality schemes will continue, and it is likely the felicitous ways of combining these will be devised. We can also hope that there will be further progress in evaluation methods. Ever since TREC, evaluation and methods to perform it have driven advances in NLP. In our view, with both datasets and evaluation, progress in the models themselves will naturally follow on.

Finally, it is certain that many papers will continue to appear. No doubt, important and relevant papers have accidentally been omitted from this review, for which we apologise in advance. You are welcome to contact the author with suggestions for papers which should be added, and we will do our best to include these in any future versions of this review.

\section*{Acknowledgments} \label{acknowledgements}
Many thanks to George Kour for the ArXiv style: https://github.com/kourgeorge/arxiv-style. Thanks to Jon Chamberlain, Yunfei Long, and Ravi Shekhar at Essex for their help and encouragement.

\bibliographystyle{abbrv}

\end{document}